# Structured Debate Improves Corporate Credit Reasoning in Financial AI

**Yoonjin Lee[1], Munhee Kim[2], Hanbi Choi[3], Juhyeon Park[4], Seungho Lyoo[5], Woojin Park[1†]**

[1]Seoul National University, [2]Openmade Consulting, [3]University of Seoul, [4]LG CNS, [5]HonestAI

yoonjin@snu.ac.kr, mhkim@openmade.co.kr, eye1719@uos.ac.kr, jparkax@lgcns.com, seungho.lyoo@honestai.tech, woojinpark@snu.ac.kr

## Abstract

This study investigated LLM–based automation for analyzing non-financial data in corporate credit evaluation. Two systems were developed and compared: a Single-Agent System (SAS), in which one LLM agent infers favorable and adverse repayment signals, and a Popperian Multi-Agent Debate System (PMADS), which structures the dual-perspective analysis as adversarial argumentation under the Karl Popper Debate protocol. Evaluation addressed three fronts: (i) work productivity (task completion time) compared with human experts; (ii) perceived report quality and usability, rated by credit risk professionals for system-generated reports; and (iii) reasoning characteristics quantified via reasoning-tree analysis (breadth, depth, and Reasoning Elaboration Index). Both systems drastically reduced task completion time relative to human experts (baseline ≈ 1,900 s/case; SAS ≈ 11.6 s; PMADS ≈ 92.0 s). Professionals rated SAS reports as adequate, while PMADS reports exceeded neutral benchmarks and scored significantly higher in explanatory adequacy, practical applicability, and usability. Reasoning-tree analysis showed PMADS produced deeper, more elaborated structures, whereas SAS yielded single-layered trees. These findings suggest that structured multi-agent debate enhances analytical rigor and perceived usefulness, though at the cost of longer computation time. Overall, the results demonstrate that reasoning-centered automation represents a promising approach for developing useful AI systems in decision-critical financial contexts.

## Introduction

The development of credit risk assessment systems has long been a central topic in finance research. A particularly important subdomain of credit risk assessment concerns the integration of non-financial data, such as governance quality, and business environment, to complement traditional financial factors in evaluating corporate repayment capacity. Such non-financial data reveal aspects of a firm's repayment capacity that conventional financial metrics alone cannot capture, thereby enhancing the robustness of credit risk assessments, especially for small and medium-sized enterprises (SMEs) (Grunert et al., 2005; Bitetto et al., 2023; Roy & Shaw, 2021; Wahlstrom et al., 2024).

In practice, non-financial data analysis is typically conducted by relationship managers (RMs) or other financial experts who synthesize information from diverse sources, weigh both favorable and adverse signals, and prepare comprehensive credit risk assessment reports on a target firm's repayment capacity. This analytical process is cognitively demanding, as analysts must interpret non-financial factors whose connections to financial outcomes are often implicit or ambiguous. Such demands can slow the assessment process and increase reliance on heuristic shortcuts, potentially compromising the accuracy and consistency of credit evaluations.

Automating the non-financial data analysis and reporting task could therefore yield substantial improvements in corporate loan service workflows if automation can achieve analytical consistency while preserving the interpretability and evidential reasoning of human analysis. Yet, research on developing such automated analytical systems remains scarce.

Large Language Models (LLMs) present a promising opportunity for automating non-financial data analysis, as they can collect, integrate, and reason over unstructured information to generate coherent, human-readable analyses (Wilson et al., 2024; Li et al., 2025). Indeed, LLMs have demonstrated effectiveness in analytical reasoning and report generation across various financial subdomains, including financial engineering, forecasting, and real-time question answering (Li et al., 2023; Nie et al., 2024; Zhao et al., 2024; Dubey et al., 2025). However, to our knowledge, no prior work has developed an LLM-powered system that analyzes non-financial data to support the evaluation of corporate credit repayment capacity. A recent study by Huang et al. (2025) utilized thirty-eight multidimensional non-financial features to predict default risk in SMEs' commercial bills and demonstrated improved predictive performance using a prompt-based LLM system. Nevertheless, their approach treated non-financial information merely as input variables for default classification and did not generate explanatory reasoning linking such information to repayment capacity.

† Corresponding Author

The long-term goal of our research is to automate corporate credit risk assessment to enhance the efficiency and quality of corporate loan services. As an initial step toward this goal, the present study developed and evaluated two LLM-powered agentic systems designed to automate non-financial data analysis and report generation. Specifically, two systems were implemented: a Single-Agent System (SAS), in which a single LLM agent identifies both favorable and adverse signals from non-financial data, and a multi-agent system that conducts the dual-perspective analyses through structured, evidence-based debates guided by a predefined debate protocol—namely, the Karl Popper Debate (KPD) protocol. Debate-based reasoning can facilitate iterative reasoning, critical reflection, and richer analytical coverage by allowing agents to challenge and refine one another's arguments; recent research on multi-agent systems has highlighted their potential (Liu, 2025; Chun et al., 2025; Fatemi, 2024; Cai et al., 2025). The KPD in particular can serve as a strong foundation for structured financial reasoning. Rooted in Popper's principle of critical rationalism—where knowledge advances through cycles of refutation and counter-refutation—the KPD emphasizes disciplined reasoning, evidence-based argumentation, and critical dialogue.

The two automation systems, the SAS and the Popperian Multi-Agent Debate System (PMADS), were empirically evaluated in terms of work productivity, perceived report quality, usability, and reasoning characteristics. We hypothesized that (1) both systems would outperform human domain experts in work productivity measured by task completion time, (2) both systems would generate data analysis results perceived as adequate by human domain experts, and (3) the PMADS would outperform the SAS across all evaluation measures.

## System Architecture and Development

### Single-Agent System (SAS)

In the SAS, when provided with a target company, the system's single LLM agent (referred to as the SAS agent) accesses company-specific data stored in a pre-existing database. This database is assumed to be continuously maintained and updated by an independent data-gathering agent that consolidates information through web crawling and API retrieval from multiple verified sources, including the Electronic Disclosure System as well as official statistical and industry policy repositories.

Once the company data are retrieved, the SAS agent extracts relevant non-financial information and conducts web searches to supplement the database with recent information when existing data are insufficient or when additional verification is required (He et al., 2023). The agent then performs a repayment capacity analysis that adopts a dual-perspective approach, integrating both favorable and adverse aspects to derive a balanced evaluation. The overall architecture of the SAS is illustrated in Figure 1, which depicts the data and reasoning flow from information collection to structured report generation.

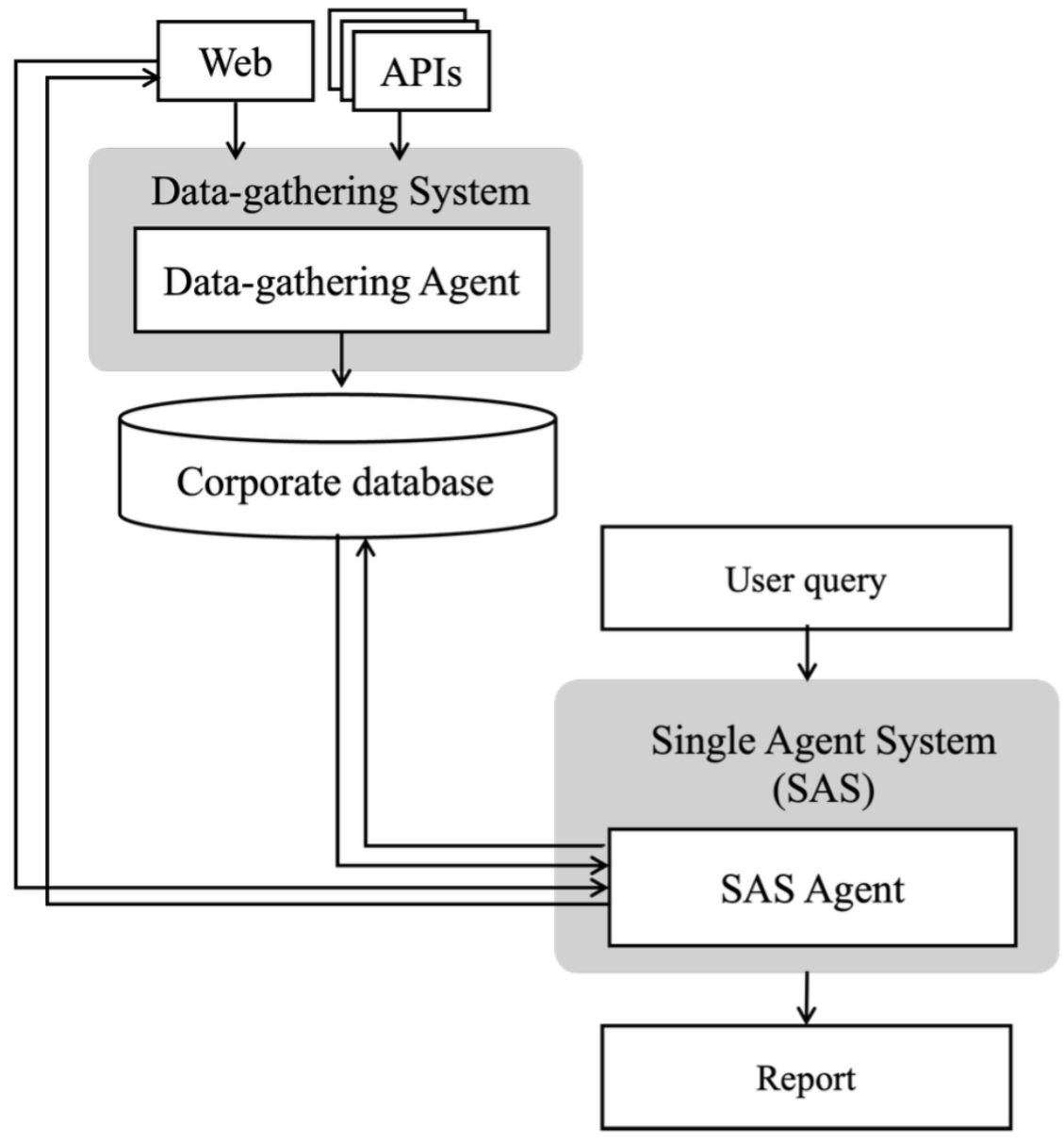

Figure 1: System architecture of SAS. Arrows indicate the directional flow of data and reasoning: corporate information is consolidated from the web and API sources into a database, then the SAS agent extracts non-financial data, performs analysis, and generates a structured report.

The SAS workflow—spanning data retrieval and extraction through reasoning and report synthesis—is guided by a predefined evaluation guideline that delineates the relevant non-financial factors and specifies how each should be interpreted in relation to repayment capacity. The key components of this guideline are summarized in Figure 2. The factors, derived from a synthesis of the credit-risk literature and industry credit-evaluation manuals used in professional practice (Grunert et al., 2005; Fu et al., 2020; Roy & Shaw, 2021; Lerner & Seru, 2022; Erdinç, 2023; Kim & Nam, 2023; Morales-Solis et al., 2023; Wahlstrøm et al., 2024; Bitetto et al., 2023; Wang et al., 2025; Haeri et al., 2025; KIS, n.d.), include industry growth outlook, industry com-petition intensity, technological change, economic sensitivity, government support, internal control risk, managerial continuity, employment stability, certification status, and search-volume trends. For each factor, the guideline defines favorable and adverse signals and explains their implications for repayment capacity.

Incorporated into the prompt used by the agent, this guideline directs the selection and interpretation of relevant evidence and the construction of claim–evidence–implication chains for each factor. The guideline-driven reasoning process is operationalized as Algorithm 1. The agent then generates a structured report for human users that includes both affirmative and adverse assessments, each supported by traceable evidence citations consistent with the guideline. The final outputs comprise an objective statement, corporate overview, synthesized affirmative and negative analyses by factor category.

| ***Non-financial factor*** | ***Interpretation criteria*** |
|---|---|
| Industry growth outlook | Higher, favorable |
| Intensity of industry competition | Stronger, adverse |
| Impact of technological change | Greater, adverse |
| Economic sensitivity | More sensitive, adverse |
| Government support programs | If present, favorable |
| Governance & compliance risk | Higher, adverse |
| Management continuity | Higher, adverse |
| Workforce stability | Higher, favorable |
| Technology & innovation certifications | If present, favorable |
| Search volume trends | Increasing, favorable |

Figure 2: Summary of the predefined evaluation guideline used in the SAS. The guideline specifies ten non-financial factors, the corresponding favorable and adverse signals, and their implications for corporate repayment capacity. In the table, *favorable* denotes signals indicative of improved repayment capacity, whereas *adverse* denotes signals indicative of diminished repayment capacity.

Algorithm 1: SAS operation logic

**Input:** K (non-financial information from corporate database), I (company identifier)
**Parameter:** R (recency window)
**Output:** F (analytical report)
1: **S** ← Summarize company using *K* and *I*.
2: **W** ← Perform web search on recent information (*I, R*).
3: **A** ← Assess the target company's loan repayment capacity based on the evaluation guideline (*S, W*).
4: **F** ← Compose a structured analytical report (*A*).

The data gathering agent and the SAS agent were implemented using the OpenAI Chat Completions API with the gpt-4o model to perform analytical reasoning and generate structured outputs. Web retrieval was conducted via SerpAPI with date-bounded constraints to ensure evidence recency and citation traceability. The generated analytical reports were logged in JSON format and subsequently parsed for evaluation. The complete SAS prompt specification is provided in Appendix A.

## Popperian Multi-Agent Debate System (PMADS)

The PMADS introduces a multi-agent architecture that formalizes adversarial verification through structured debate. The system consists of two coordinated subsystems: a debate subsystem and an aggregator subsystem. Its overall architecture is illustrated in Figure 3. In the debate subsystem, six LLM agents conduct a repayment capacity analysis for a given company following the KPD protocol. The agents are assigned complementary discourse roles and organized into affirmative (A1–A3) and negative (N1–N3) teams, with each team arguing that the target firm's repayment capability is either favorable or at risk. In the aggregator subsystem, a single LLM agent synthesizes the debate outcomes into a final structured report.

The debate proceeds according to the KPD protocol, a ten-step sequence designed to iteratively strengthen, challenge, and refine competing claims. The protocol is summarized as follows:

1. A1 introduces affirmative constructive of the repayment supportive claim based on favorable signals.
2. N3 cross-examines the affirmative constructive.
3. N1 introduces negative constructive of the repayment risk claim based on adverse signals.
4. A3 cross-examines the negative constructive.
5. A2 rebuttal refutes N1's negative constructive.
6. N1 cross-examines A2's rebuttal.
7. N2 rebuttal refutes A1's affirmative constructive.
8. A1 cross-examines N2's rebuttal.
9. A3 synthesizes affirmative constructive, rebuttal, and cross-examinations from negative team and composes the final consolidated arguments.
10. N3 synthesizes negative constructive, rebuttal, and cross-examinations from affirmative team and composes the final consolidated arguments.

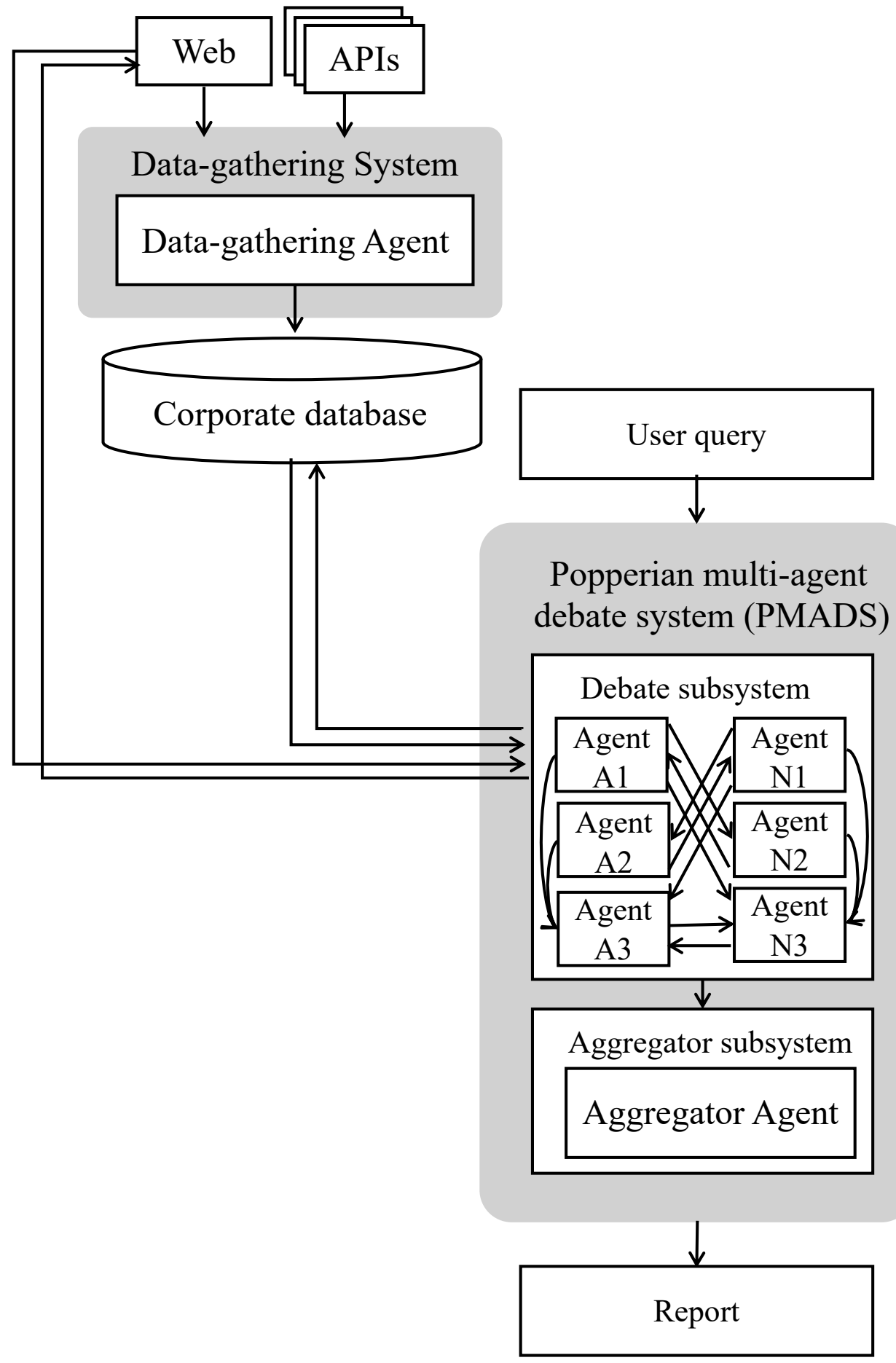

Figure 3: System architecture of PMADS. Arrows indicate the directional flow of data and reasoning: corporate information is consolidated from the web and API sources into a database, then the PMADS debate subsystem retrieves relevant non-financial data and reasons according to the KPD protocol; the aggregator subsystem synthesizes the arguments into a structured report.

Each debate step constitutes a discrete task assigned to a specific agent. Within each task, the corresponding agent performs role-specific reasoning using the available data while ensuring explicit evidence citation and factor-relevance tagging. To prevent discourse drift, only essential context is passed between steps to the agent responsible for the subsequent task. The final affirmative and negative assessments are produced in Steps 9 and 10, where both teams integrate the accumulated debate content into their respective closing statements. An illustrative example of discourse propagation leading to the construction of the final affirmative closing argument (Step 9) is provided in Appendix B; the opposing team follows a symmetric procedure. The overall operational logic implementing the ten debate steps is formalized in Algorithm 2, which specifies agent instantiation, retrieval policies, and role-specific reasoning functions.

---

Algorithm 2: PMADS operation logic

---

**Input**: K (non-financial information from corporate database, I (company identifier)
**Parameters**: R (recency window), U (number of non-financial factors to be considered), C (maximum number of characters), D (maximum number of characters), Q (number of cross-examination questions)
**Output**: F (analytical report)

1: **S** ← Summarize company using *K* and *I*.
2: **Debate step 1 (Affirmative constructive, A1):**
  **D1** ← A1 assesses the target company in the perspective that loan repayment capacity will improve, using at least *U* favorable non-financial factors based on the evaluation guideline (*S*). If required, recent information is retrieved via web search (*I, R*). Reasoning is required to be falsifiable. Output length is limited to *C* characters.
3: **Debate step 2 (Cross-examination, N3):**
  **D2** ← N3 composes *Q* cross-examination questions targeting weakness in the affirmative constructive, based on the evaluation guideline (*D1, S*).
4: **Debate step 3 (Negative constructive, N1):**
  **D3** ← N1 assesses the target company from the perspective that loan repayment capacity will weaken or is at risk, using at least *U* adverse non-financial factors based on the evaluation guideline (*S*). If required, recent information is retrieved via web search (*I, R*). Reasoning is required to be falsifiable. Output length is limited to *C* characters.
5: **Debate step 4 (Cross-examination, A3):**
  **D4** ← A3 generates *Q* cross-examination questions targeting weakness in the negative constructive, based on the evaluation guideline (*D3, S*).
6: **Debate step 5 (Rebuttal from affirmative team, A2):**
  **D5** ← A2 rebuts the negative constructive using counter-interpretation and supporting evidence, based on the evaluation guideline (*D3, S*). If required, recent information is retrieved via web search (*I, R*). Output length is limited to *D* characters.
7: **Debate step 6 (Cross-examination, N1):**
  **D6** ← N1 generates *Q* cross-examination questions targeting weaknesses in the affirmative rebuttal, based on the evaluation guideline (*D5, S*).
8: **Debate step 7 (Rebuttal from negative team, N2):**
  **D7** ← N2 rebuts the affirmative constructive by identifying risks, uncertainties, or negative implications based on the evaluation guideline (*D1, S*). If required, recent information is retrieved via web search (*I, R*). Output length is limited to *D* characters.
9: **Debate step 8 (Cross-examination, A1):**

---

**D8** ← A1 generates *Q* cross-examination questions targeting weakness in the negative rebuttal, based on the evaluation guideline (*D7, S*).

10: **Debate step 9 (Affirmative summary, A3):**
**D9** ← A3 summarizes the affirmative constructive, rebuttal, and cross-examinations from the negative team into a closing summary within *C* characters, based on the evaluation guideline (*D1, D2, D5, D8, S*).

11: **Debate step 10 (Negative summary, N3):**
**D10** ← N3 summarizes the negative constructive, rebuttal, and cross-examinations from the affirmative team into a closing summary within *C* characters, based on the evaluation guideline (*D3, D4, D6, D7, S*).

12: **F** ← The aggregator agent composes a structured analytical report (*D1, … , D10*).

Following the debate, the aggregator subsystem integrates the arguments into a structured report for human users. The report presents both affirmative and adverse assessments, each supported by traceable evidence citations consistent with the evaluation guideline. Rather than selecting a "winner," the report preserves both perspectives to facilitate expert decision-making. The final outputs include an objective summary statement, a corporate overview, synthesized affirmative and negative analyses by factor category, and the full debate transcript to ensure audit transparency.

The PMADS was implemented on the CrewAI framework (v0.148.0), with all six agents instantiated as GPT-4o models via the OpenAI API. CrewAI managed scheduling and controlled context transfer across debate turns. Web searches were executed through SerpAPI under the same recency constraints as the SAS, and all debate transcripts and analytical reports were logged in JSON for evaluation. The complete PMADS prompt specification is presented in Appendix C.

## System evaluation

### Methods

To examine the usefulness of the two systems in a realistic credit analysis context, a case-based evaluation was conducted using three South Korean companies. Each company exhibited a heterogeneous profile comprising favorable, adverse, and context-dependent non-financial factors relevant to credit risk. In total, six reports (3 companies × 2 systems) were generated and evaluated.

The two systems and human domain experts were compared in terms of work productivity, measured by task completion time. Baseline estimates of manual end-to-end report completion times per company were obtained from a practitioner survey, and the mean baseline value served as a reference for quantifying productivity improvement achieved by each system. For each of the three companies (A, B, and C), average end-to-end processing times were recorded for both systems.

The two systems were also evaluated using additional subjective and objective measures. Subjective evaluation focused on perceived report quality and usability. Five industry credit risk professionals (three relationship managers and two credit specialists) participated in the evaluation. Each participant rated every report on three criteria tailored to the context of system use—trustworthiness ("the contents of the report are logically coherent and can be regarded as reliable"), explanatory adequacy ("the report presents adequately reasoned support for evaluating repayment capacity"), and practical applicability ("the evaluation of company repayment capacity in the report is applicable in practice as an aid to decision-making")—as well as on usability, measured using the System Usability Scale (SUS). All ratings were collected on a 5-point Likert scale (1 = strongly disagree, 5 = strongly agree). To prevent bias, participants were not informed of the source (SAS or PMADS) of the six reports. The presentation order of the reports was randomized to control for order effects. Paired scores from the same participants were compared using Wilcoxon signed-rank tests to examine differences between the two systems for each company.

Objective evaluation focused on the reasoning characteristics of system-generated reports. Each report's reasoning structure was first represented as a hierarchical reasoning tree, with the final claim as the root node, supporting arguments and counterarguments as subordinate nodes, and logical relations as connecting branches. From these reasoning trees, three metrics were derived to quantify how extensively and hierarchically each system developed its reasoning: breadth, depth, and the Reasoning Elaboration Index (REI). Breadth was defined as the number of distinct topical branches supporting the root claim, and depth as the number of hierarchical argumentation levels within each branch. The REI was formulated to integrate these two dimensions, reflecting cumulative elaboration across topics and rewarding multi-level argument development even when breadth was limited. Specifically, REI was calculated as:

$$\text{REI} = \sum_{t \in T} depth_t$$

where $t$ is the set of distinct topical branches supporting the root claim.

Reasoning trees were generated using GPT-4o and subsequently verified by two researchers. Final REI scores were averaged across reports for system-level comparison. An illustrative example of the reasoning-tree construction and REI derivation process is presented in Appendix D.

### Evaluation Results

The two systems were found to significantly reduce the time required to generate a report per company. The credit risk professionals reported that they typically spent an average of 1,900.2 seconds per case analyzing approximately 14 companies per week. In comparison, the SAS required an average of 11.55 seconds per case, while the PMADS required an average of 91.97 seconds per case. Table 1 summarizes the task completion time data for the three companies.

| System | A | B | C |
|---|---|---|---|
| SAS | 10.33 | 12.51 | 11.81 |
| PMADS | 75.02 | 98.62 | 102.27 |

Table 1: Task completion times (sec) for companies A, B, and C using SAS and PMADS.

Wilcoxon signed-rank tests were conducted on fifteen paired observations (3 companies × 5 experts) comparing the SAS and PMADS across the four subjective measures: trustworthiness, explanatory adequacy, practical applicability, and usability. Figure 4 presents the median scores for the two systems on each measure. Trustworthiness ratings did not differ significantly between the systems, W = 10.5, Z = –1.422, p = .141, r = –0.474 (n = 9; $\text{Mdn}_{SAS}$ = 3.0, $\text{Mdn}_{PMADS}$ = 4.0). Explanatory adequacy ratings were significantly higher for the PMADS, W = 0.0, Z = –3.059, p = .002, r = –0.883 (n = 12; $\text{Mdn}_{SAS}$ = 3.0, $\text{Mdn}_{PMADS}$ = 4.0). Practical applicability ratings also showed a significant difference favoring the PMADS, W = 0.0, Z = –3.059, p = .002, r = –0.883 (n = 12; $\text{Mdn}_{SAS}$ = 3.0, $\text{Mdn}_{PMADS}$ = 4.0). SUS scores were significantly higher for the PMADS, W = 9.0, Z = –2.353, p = .018, r = –0.679 (n = 12; $\text{Mdn}_{SAS}$ = 52.5, $\text{Mdn}_{PMADS}$ = 62.5). The effective sample size (n) varied across tests because pairs with zero difference were excluded from the analysis, as required by the Wilcoxon procedure.

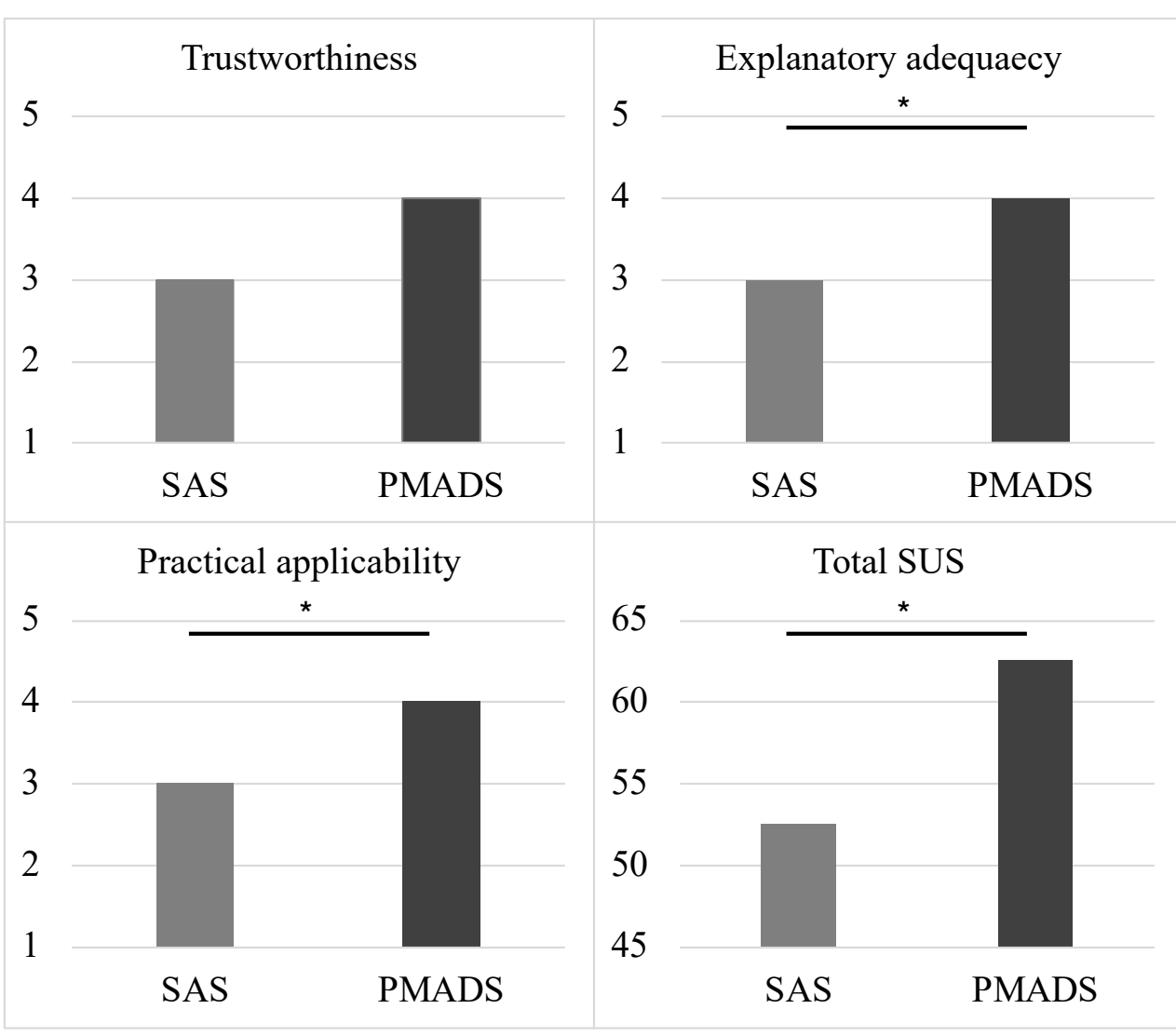

Figure 4: Median subjective evaluation scores comparing the SAS and the PMADS.

Regarding the objective evaluation, Table 2 presents the results of the reasoning tree analysis comparing the PMADS and the SAS in terms of breadth, depth, and the Reasoning Elaboration Index (REI). On average, the REI for the PMADS (M = 14.33 ± 3.21) exceeded that of the SAS (M = 8.00 ± 1.00). The difference in REI was primarily driven by greater depth in the PMADS. The SAS produced broader reasoning trees (breadth = 7–9) with a single-level depth across all companies, whereas the PMADS generated slightly narrower but deeper trees (breadth = 6–7; depth up to three levels).

| Metric | System | A | B | C |
|---|---|---|---|---|
| Breadth | SAS | 7 | 8 | 9 |
| | PMADS | 6 | 7 | 6 |
| Depth | SAS | 1 | 1 | 1 |
| | PMADS | 1, 3 | 1, 3 | 3 |
| REI | SAS | 7 | 8 | 9 |
| | PMADS | 12 | 13 | 18 |

Table 2: Results of reasoning tree analysis comparing the SAS and the PMADS in terms of breadth, depth, and Reasoning Elaboration Index (REI) across companies A, B, and C.

Notably, the depth of reasoning in the PMADS varied across non-financial factors. In Company A, three of six factors reached a depth of three levels; in Company B, three of seven factors reached that level; and in Company C, all six factors reached three levels, while the remaining non-financial factors in Companies A and B remained at a single level.

## Discussion and Future Works

This study evaluated two LLM-based automation systems—the Single-Agent System (SAS) and the Popperian Multi-Agent Debate System (PMADS)—for non-financial data analysis in corporate credit evaluation. The findings supported all three hypotheses. First, both systems substantially outperformed human domain experts in work productivity, as reflected by dramatically shorter task completion times. Second, reports generated by the SAS were perceived by credit risk professionals as adequate or neutral in quality, while those produced by the PMADS were perceived as more than adequate. Third, the PMADS was superior to the SAS across most perceived report quality measures, including explanatory adequacy, practical applicability, and usability, while trustworthiness ratings remained comparable between the two systems.

The combination of neutral or higher perceived quality and drastically reduced task completion time highlights the practical utility of LLM-powered automation for corporate credit assessment. Both systems demonstrated the potential to alleviate the cognitive and temporal burdens associated with manual report generation, indicating that automation can complement expert judgment rather than replace it. The PMADS, in particular, showed promise as a viable analytical assistant in credit evaluation, as its perceived report quality and usability exceeded neutral benchmarks, suggesting that users found its reasoning coherent and its interface operationally acceptable for real-world workflows.

The superior reasoning quality of the PMADS over the SAS can be attributed to its debate protocol, which formalizes analysis as a structured process of argumentation. This design accords with the *argumentative theory of reasoning* (Mercier & Sperber, 2011), which posits that reasoning quality improves when claims are subject to critique rather than produced in isolation. By organizing analysis into sequential stages that require claims to be evidenced, evaluated, and revised, the protocol filters unsupported assumptions and prevents premature closure during inference. Consistent with recent multi-agent research, structured critique and procedural constraints have been shown to enhance reasoning stability and factual reliability by mitigating coordination failures and conversational drift (Gao et al., 2025; Jin et al., 2025). Likewise, Sreedhar and Chilton (2025) demonstrated in a strategic negotiation task that multi-agent interaction—requiring agents to iteratively justify and adjust their positions—yields more coherent and defensible reasoning than single-agent generation.

The PMADS exhibited more elaborate reasoning than the SAS, as reflected by its higher Reasoning Elaboration Index (REI) and greater reasoning depth. While the SAS generated broader but single-layered reasoning trees, the PMADS produced narrower yet deeper trees, indicating a more hierarchical and reflective reasoning structure. Notably, the depth of reasoning varied across non-financial factors: in Company A, three of six factors reached a depth of three levels; in Company B, three of seven factors reached that level; and in Company C, all six factors reached three levels, while the remaining factors in Companies A and B remained at a single level. These findings suggest that structured debate encourages deeper reasoning when factor-related evidence is sufficiently rich or contentious.

A clear trade-off was observed between reasoning quality and task completion time. While the PMADS produced reports with more elaborate and defensible reasoning, its analysis required substantially longer execution time compared with the SAS. This trade-off underscores an important design consideration for real-world applications: optimizing the balance between reasoning depth and processing efficiency. In practice, system selection may depend on contextual demands—whether rapid report generation or in-depth analytical justification is prioritized.

Taken together, the results demonstrate that LLM-powered automation can effectively support non-financial data analysis for evaluating corporate repayment capacity. Both systems achieved substantial productivity gains while maintaining acceptable report quality, and the PMADS, in particular, exhibited superior reasoning performance and user acceptance. These findings affirm the feasibility of reasoning-centered AI systems as assistive tools for credit risk professionals and suggest that structured multi-agent reasoning can meaningfully enhance interpretability and analytical rigor in financial decision-support contexts.

Future research may extend this work in four directions. First, debate protocol design should be systematically examined by varying procedural parameters such as turn-allocation strategy, rebuttal depth, and challenge frequency to optimize the balance between rigor and computational efficiency. Second, architectural enhancement through agent specialization could further increase reasoning coherence by assigning differentiated analytical roles, adopting alternative debate protocols, or embedding domain expertise. Third, the applicability of the PMADS framework may be explored beyond corporate credit risk assessment, particularly in high-stakes financial environments such as insurance underwriting and project finance evaluation. Advancing these directions will deepen understanding of how structured reasoning can be engineered for defensible, decision-critical AI.

Finally, in inference-heavy settings like credit evaluation—where decisions must be made under ambiguous and incomplete signals—one important failure mode is reasoning-level hallucination: conclusions that are not supported by the available evidence due to fallacy-prone interpretations of otherwise factual inputs. Notably, hallucinations can also stem from ungrounded factual claims (e.g., fabricated details or sources); the focus here is on the subset driven by invalid inference from ambiguous signals. In such cases, improving retrieval and factual grounding may be insufficient; systems must also reduce systematic inference errors. Hong et al. (2024) provided a taxonomy and prompting strategies for fallacy detection in foundation-model outputs. A promising direction is to equip agentic systems with explicit fallacy knowledge and mechanisms that detect fallacious reasoning and trigger re-inference. Future work should therefore focus on operationalizing fallacy knowledge in machine-interpretable forms to enable self-checking and iterative correction of reasoning errors, complementing factual grounding.

## Declaration of generative AI and AI-assisted technologies in the manuscript preparation process

During the preparation of this work the author(s) used ChatGPT in order to refine English expression of the manuscript. After using this tool/service, the author(s) reviewed and edited the content as needed and take(s) full responsibility for the content of the published article.

## Reference

Dubey, S. S.; Astvansh, V.; and Kopalle, P. K. 2025. Generative AI solutions to empower financial firms. *Journal of Public Policy & Marketing* 07439156241311300.

Gao, M.; Li, Y.; Liu, B.; Yu, Y.; Wang, P.; Lin, C. Y.; and Lai, F. 2025. Single-agent or multi-agent systems? Why not both? arXiv:2505.18286.

Haeri, A.; Vitrano, J.; and Ghelichi, M. 2025. Generative AI enhanced financial risk management information retrieval. arXiv:2504.06293.

Huang, H.; Li, J.; Zheng, C.; Chen, S.; Wang, X.; and Chen, X. 2025. Advanced default risk prediction in small and medium-sized enterprises using large language models. *Applied Sciences* 15(5): 2733.

Jin, C.; Peng, H.; Zhang, Q.; Tang, Y.; Metaxas, D. N.; and Che, T. 2025. Two heads are better than one: Test-time scaling of multi-agent collaborative reasoning. arXiv:2504.09772.

Li, H.; Gao, H.; Wu, C.; and Vasarhelyi, M. A. 2025. Extracting financial data from unstructured sources: Leveraging large language models. *Journal of Information Systems* 39(1): 135-156.

Sreedhar, K.; and Chilton, L. 2025. Simulating strategic reasoning: Comparing the ability of single LLMs and multi-agent systems to replicate human behavior.

Wang, L.; Verousis, T.; and Zhang, M. 2025. Market value of R&D, patents, and CEO characteristics. *Financial Innovation* 11(1): 8.

Nie, Y.; Kong, Y.; Dong, X.; Mulvey, J. M.; Poor, H. V.; Wen, Q.; and Zohren, S. 2024. A survey of large language models for financial applications: Progress, prospects and challenges. arXiv:2406.11903.

Wahlstrøm, R. R.; Becker, L. K.; and Fornes, T. N. 2024. Enhancing credit risk assessments of SMEs with non-financial information. *Cogent Economics & Finance* 12(1): 2418910.

Wilson, E.; Saxena, A.; Mahajan, J.; Panikulangara, L.; Kulkarni, S.; and Jain, P. 2024. FIN2SUM: Advancing AI-driven financial text summarization with LLMs. In *2024 International Conference on Trends in Quantum Computing and Emerging Business Technologies*, 1-5. IEEE.

Zhao, H.; Liu, Z.; Wu, Z.; Li, Y.; Yang, T.; Shu, P.; and Liu, T. 2024. Revolutionizing finance with LLMs: An overview of applications and insights. arXiv:2401.11641.

Bitetto, A.; Cerchiello, P.; Filomeni, S.; Tanda, A.; and Tarantino, B. 2023. Machine learning and credit risk: Empirical evidence from small-and mid-sized businesses. *Socio-Economic Planning Sciences* 90: 101746.

Çetin, A. İ.; Çetin, A. E.; and Ahmed, S. E. 2023. The impact of non-financial and financial variables on credit decisions for service companies in Turkey. *Journal of Risk and Financial Management* 16(11): 487.

Feng, D.; Dai, Y.; Huang, J.; Zhang, Y.; Xie, Q.; Han, W.; and Wang, H. 2023. Empowering many, biasing a few: Generalist credit scoring through large language models. arXiv:2310.00566.

He, H.; Zhang, H.; and Roth, D. 2023. Rethinking with retrieval: Faithful large language model inference. CoRR abs/2301.00303.

Kim, N.; and Nam, J. 2023. 빅데이터 분석을 활용한 중소기업·소상공인 ESG 경영 이슈 분석: 시기별·기업 규모별 비교를 중심으로. *Entrepreneurship & ESG 연구* 3(2): 1-28.

Li, Y.; Wang, S.; Ding, H.; and Chen, H. 2023. Large language models in finance: A survey. In *Proceedings of the Fourth ACM International Conference on AI in Finance*, 374-382. New York: Association for Computing Machinery.

Morales-Solis, J. C.; Barker III, V. L.; and Cordero, A. M. 2023. CEO's industry experience and emerging market SME performance: The effects of corruption and political uncertainty. *Journal of Business Venturing Insights* 20: e00424.

Yoon, S. 2023. Design and implementation of an LLM system to improve response time for SMEs technology credit evaluation. *The International Journal of Advanced Smart Convergence* 12(3): 51-60.

Lerner, J.; and Seru, A. 2022. The use and misuse of patent data: Issues for finance and beyond. *The Review of Financial Studies* 35(6): 2667-2704.

Roy, P. K.; and Shaw, K. 2021. A multicriteria credit scoring model for SMEs using hybrid BWM and TOPSIS. *Financial Innovation* 7(1): 77.

Fu, G.; Sun, M.; and Xu, Q. 2020. An alternative credit scoring system in China's consumer lending market: A system based on digital footprint data. SSRN 3638710.

Mercier, H.; and Sperber, D. 2011. Why do humans reason? Arguments for an argumentative theory. *Behavioral and Brain Sciences* 34(2): 57-74.

Grunert, J.; Norden, L.; and Weber, M. 2005. The role of non-financial factors in internal credit ratings. *Journal of Banking & Finance* 29(2): 509-531.

KIS Credit Rating. n.d. 신용평가 일반론 (Rating Methodology). https://www.bond.co.kr/post/dawn/10073. Accessed: 2025-07-01.

Hong, R., Zhang, H., Pang, X., Yu, D., & Zhang, C. (2024, June). A closer look at the self-verification abilities of large language models in logical reasoning. In Proceedings of the 2024 Conference of the North American Chapter of the Association for Computational Linguistics: Human Language Technologies (Volume 1: Long Papers) (pp. 900-925).

# Appendices

Appendix A. SAS analysis prompt

**Guideline prompt**

guideline = """

* Data contained in the JSON may be used as non-financial information. Non-financial information may include news, certification records, patents, governance data, and other qualitative factors.

※ Follow the Non-Financial Evaluation Guideline below. If a data element in the JSON does not fall under any of the categories listed, interpret it as (+) or (-) according to context.

[Non-Financial Evaluation Guideline]

| Non-Financial Factor | Interpretation Standard |
| Industry Growth Outlook | Higher = (+) |
| Industry Competitive Intensity | Stronger = (-) |
| Technological Disruption Risk | Higher = (-) |
| Economic Cyclicality | Higher sensitivity = (-) |
| Government Support Programs | Presence = (+) |
| Internal Control Risk | Higher = (-) |
| Management Continuity | More stable = (+) |
| Employment Stability | Higher = (+) |
| Certifications (e.g., INNOBIZ) | Presence = (+) |
| Search Trend Volume | Increasing = (+) |

※ (+): favorable signal for loan repayment capacity / (-): adverse signal

※ Do not repeatedly use the same non-financial factor across arguments from either side.

※ Each statement must include distinct non-financial factors. Use a balanced variety of data from the company_data JSON (news, governance, patents, certifications, etc.), not only company_summary.

※ The same factor may be reused only if supported by clearly different evidence (e.g., different years, different values, or separate events), and such distinctions must be explicitly described.

[Principle: Priority on Most Recent Information]

※ If multiple records of the same factor exist in the JSON, always use the most recent data first.

※ When data include dates (e.g., from news, certifications, patents), cite them in descending chronological order and prioritize the latest information.

※ If news coverage is limited or missing, you must supplement it using SerpAPISearchTool with the latest available information.

[Case 1 – News Supplement] Search Keyword: '{Company Name} + latest news'

※ If data for rebuttal or argumentation is insufficient in the JSON, you must supplement it using SerpAPISearchTool.

[Case 2 – Argument Support] Search Keyword: '{Company Name} + {specific topic} + latest trends'

※ When using past data, always specify the date clearly (e.g., "as of March 2025") and interpret it in the context of current trends.

"""

**Analysis prompt**

You are a corporate credit evaluation expert. Based on the given non-financial data and the latest web search results, you must generate an analytical report assessing a company's loan repayment capacity.

- You must strictly follow the JSON output format shown below.
- Each element (positive and negative factors) must be written as richly as possible, including concrete examples, figures, citations, and sources.

- For each topic, clearly distinguish between supportive argument and adverse argument) and include the source of information (web search, company data, etc.) in every argument.
- All analytical evidence must rely solely on the [Company Summary] and the [Latest Web Search Results] provided below. Assumptions, fabrication, or direct inference beyond the given data is strictly prohibited.

[Example JSON Output]

```
{
  "Analysis Summary": {
    "Favorable Factors Summary": [
      "Potential mitigation of risk through supply chain diversification",
      "Foundation for capital attraction strengthened by ESG strategy initiatives"
    ],
    "Adverse Factors Summary": [
      "Uncertainty regarding ESG contribution to short-term profitability",
      "Declining competitiveness due to insufficient response to technological change"
    ],
    "topics": [
      {
        "topic": "Supply Chain Stability",
        "Affirmative": "Pursuit of diversification strategy → potential for risk mitigation. Source: 'Ministry of Trade, Supply Chain Stability Briefing (2024)'. Reduced supply risk → potential maintenance of stable revenue.",
        "Adverse": "Uncertainty in alternative supplier quality/contracts. Source: 'KITA Supply Chain Brief (Jan 2024)'. If supply delays persist → risk of delivery failures → revenue decline and elevated credit risk."
      }
    ]
  }
}
```

Appendix B. Discourse flow diagram of the PMADS showing the sequential development of arguments and context transitions among agents throughout the debate turns

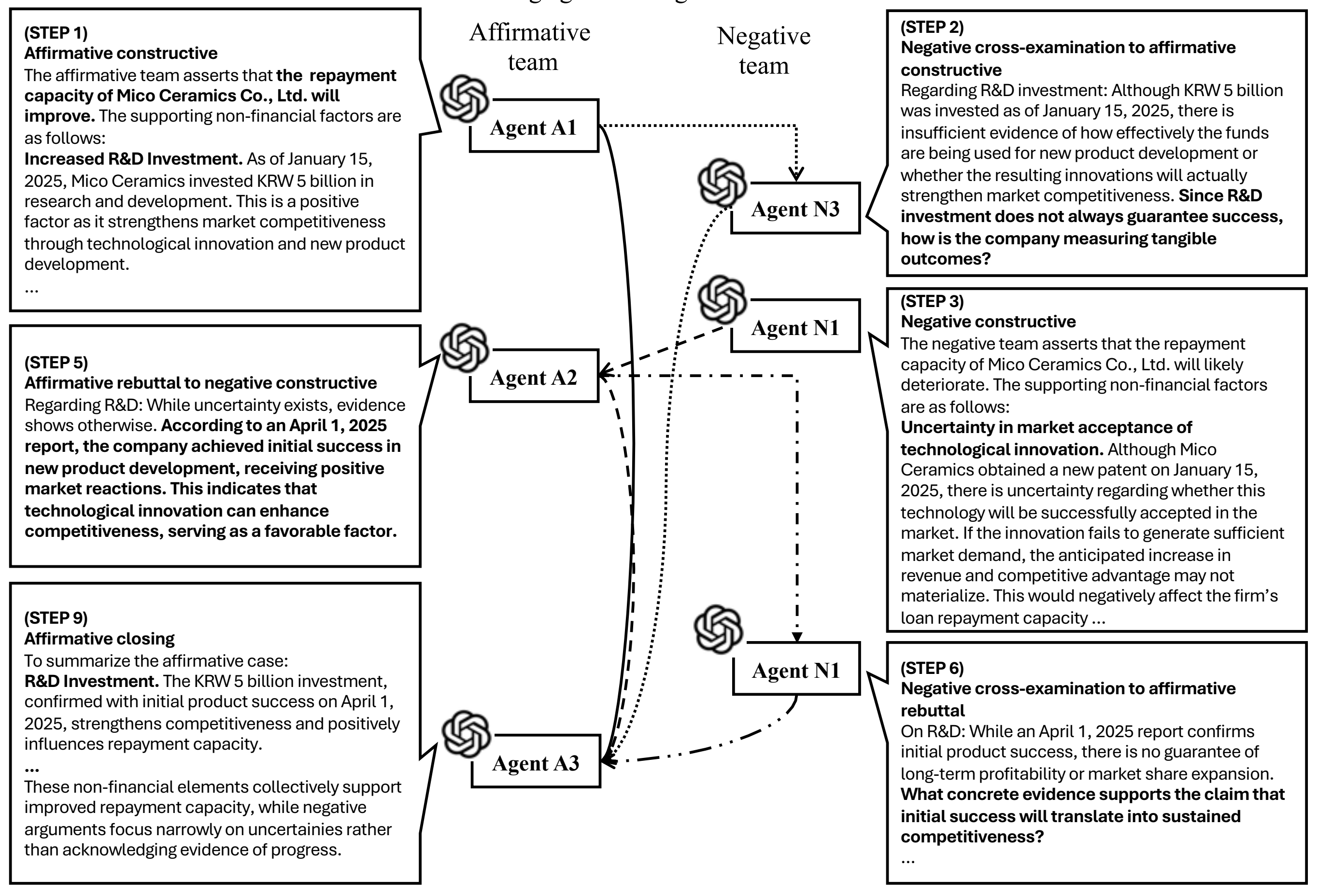

Appendix C. PMADS analysis prompt in CrewAI framework

**Guideline prompt**

guideline = """

* Data contained in the JSON may be used as non-financial information. Non-financial information may include news, certification records, patents, governance data, and other qualitative factors.

※ Follow the Non-Financial Evaluation Guideline below. If a data element in the JSON does not fall under any of the categories listed, interpret it as (+) or (-) according to context.

[Non-Financial Evaluation Guideline]

| Non-Financial Factor | Interpretation Standard |
|---|---|
| Industry Growth Outlook | Higher = (+) |
| Industry Competitive Intensity | Stronger = (-) |
| Technological Disruption Risk | Higher = (-) |
| Economic Cyclicality | Higher sensitivity = (-) |
| Government Support Programs | Presence = (+) |
| Internal Control Risk | Higher = (-) |
| Management Continuity | More stable = (+) |
| Employment Stability | Higher = (+) |
| Certifications (e.g., INNOBIZ) | Presence = (+) |

| Search Trend Volume | Increasing = (+) |

※ (+): favorable signal for loan repayment capacity / (-): adverse signal

※ Do not repeatedly use the same non-financial factor across arguments from either side.

※ Each statement must include distinct non-financial factors. Use a balanced variety of data from the company_data JSON (news, governance, patents, certifications, etc.), not only company_summary.

※ The same factor may be reused only if supported by clearly different evidence (e.g., different years, different values, or separate events), and such distinctions must be explicitly described.

※ The purpose of the Karl Popper Debate is to evaluate both risks and opportunities comprehensively; therefore, use favorable signal, adverse signal, and context-dependent impact requiring expert evaluation signals in a balanced manner.

[Principle: Priority on Most Recent Information]

※ If multiple records of the same factor exist in the JSON, always use the most recent data first.

※ When data include dates (e.g., from news, certifications, patents), cite them in descending chronological order and prioritize the latest information.

※ If news coverage is limited or missing, you must supplement it using SerpAPISearchTool with the latest available information.

[Case 1 – News Supplement] Search Keyword: '{Company Name} + latest news'

※ If data for rebuttal or argumentation is insufficient in the JSON, you must supplement it using SerpAPISearchTool.

[Case 2 – Argument Support] Search Keyword: '{Company Name} + {specific topic} + latest trends'

※ When using past data, always specify the date clearly (e.g., "as of March 2025") and interpret it in the context of current trends.

"""

**Debate process prompt**

karl_popper_explanation = """

[Debate Method: Karl Popper Debate]

This debate aims to evaluate a company's loan repayment capacity using non-financial information in order to generate a report from a financial institution's perspective. The objective is to predict the company's repayment outlook after the DART disclosure date based on non-financial evidence. Data prior to the DART disclosure date must be reviewed with caution.

This debate must follow the Karl Popper Debate structure, and comply with the following rules and procedures:

1. The debate must strictly follow the 10-step speaking structure:

A1 Constructive → N3 Cross-examination → N1 Constructive → A3 Cross-examination →A2 Rebuttal → N1 Cross-examination → N2 Rebuttal → A1 Cross-examination → A3 Final Affirmative Summary → N3 Final Negative Summary

2. Each statement must follow this logical structure:

- Present one core claim
- Support the claim using at least three non-financial factors
- All evidence must cite specific figures/dates/details explicitly from the company_data JSON, not company_summary, prioritizing the most recent data
- Include falsifiability by acknowledging uncertainty, counter-conditions, or exceptions

3. Every claim must follow the [Non-Financial Evaluation Guideline] using favorable or adverse signals. If a factor is not in the guideline, classify it as favorable or adverse signals based on context.

4. Fabrication, excessive assumptions, and creative speculation are strictly prohibited. Arguments must stay within the boundaries of the provided input data.

5. The debate must help the reader assess expected credit impact for each topic by clearly presenting implications for loan repayment capacity.

6. If not all information in the given JSON data has been discussed, the debate must continue following the 10-step format until all data is reviewed. If all information has been covered, the debate may conclude.

7. The entire debate must be written in Korean.

8. After all debate rounds are complete, both affirmative and negative teams must provide a final summarized conclusion.""""

**Agent prompt**

**A1**

role="Affirmative Speaker A1",

goal=("Present an argument that the company's loan repayment capacity will improve based on non-financial information, following the Karl Popper debate method."),

backstory=(karl_popper_explanation +

"You are an optimistic financial analyst skilled in the Karl Popper debate format."

"As the first affirmative speaker, present one core claim supported by at least three non-financial factors that contribute to improved loan repayment capacity."

"You must cite concrete data with explicit source references and include falsifiability in your reasoning. All responses must be written in Korean."

"[Case 1] If news information is insufficient: use SerpAPISearchTool with the keyword '{Company Name} + latest news'"

"※ Always prioritize the most recent information from the JSON company_data."),

tools=[websearch_tool] if websearch_tool and websearch_tool != [] else []

**A2**

role="Affirmative Rebuttal Speaker A2",

goal=(

"Refute the negative team's argument according to the Karl Popper debate method."

"[Case 2] If rebuttal information is insufficient: use SerpAPISearchTool with the keyword '{Company Name} + {specific topic} + latest trends' to find and cite credible sources."

),

backstory=(karl_popper_explanation +

"You are an optimistic financial analyst adept at structured argumentation in the Karl Popper debate format."

"Identify weaknesses in the opposing argument, including errors in interpretation, logical gaps, or uncertain data. When JSON data is insufficient, use real-time web search to supplement rebuttal evidence."

"[Case 2] If debate evidence is insufficient: use SerpAPISearchTool with the keyword '{Company Name} + {specific topic} + latest trends'"

"All responses must be written in Korean."

"※ Always prioritize the most recent information from the JSON company_data."),

tools=[websearch_tool] if websearch_tool and websearch_tool != [] else []

**A3**

role="Affirmative Summarizer and Cross-Examiner A3",

goal="Develop cross-examination questions targeting the negative team and summarize the affirmative team's arguments.",

backstory=(karl_popper_explanation +

"You are a logically rigorous analyst with strong critical thinking and summarization ability, skilled in the Karl Popper debate format."

"Develop three cross-examination questions that identify logical weaknesses in the opposing team's statements and conclude by summarizing the affirmative position persuasively."

"All responses must be written in Korean.")

**N1**

role="Negative Speaker N1",

goal="Present a counterargument that challenges the claim that loan repayment capacity will improve, following the Karl Popper debate method.",

backstory=(karl_popper_explanation +

"You are a pessimistic financial risk analyst skilled in the Karl Popper debate method. As the first negative speaker, present one core claim"

"supported by at least three non-financial factors signaling negative or uncertain impact on loan repayment capacity."

"Your argument must include data-based citations and consideration of falsifiability."

"All responses must be written in Korean."

"[Case 1] If news information is insufficient: use SerpAPISearchTool with the keyword '{Company Name} + latest news'"

"※ Always prioritize the most recent information from the JSON company_data."),

tools=[websearch_tool] if websearch_tool and websearch_tool != [] else []

**N2**

role="Negative Rebuttal Speaker N2",

goal=("Refute the affirmative team's constructive argument according to the Karl Popper debate method."

"[Case 2] If rebuttal evidence is insufficient: use SerpAPISearchTool with the keyword '{Company Name} + {specific topic} + latest trends' to find credible sources."),

backstory=(karl_popper_explanation +

"You are a pessimistic financial analyst skilled in structured rebuttal strategy within the Karl Popper debate method."

"Identify and challenge over-optimistic assumptions, data inconsistencies, and alternative interpretations."

"When JSON data is insufficient, use real-time web search to supplement factual rebuttal evidence."

"All responses must be written in Korean."

"※ Always prioritize the most recent information from the JSON company_data."),

tools=[websearch_tool] if websearch_tool and websearch_tool != [] else []

**N3**

role="Negative Summarizer and Cross-Examiner N3",

goal=("Develop cross-examination questions targeting the affirmative team and deliver the final negative summary."),

backstory=(karl_popper_explanation +

"You are a critical analyst with strong logical reasoning and summarization skills, well-versed in the Karl Popper debate method."

"Develop three cross-examination questions that expose weaknesses in the affirmative argument and conclude by summarizing the negative position persuasively."

"All responses must be written in Korean.")

**Aggregator**

role="Aggregator Agent",

goal=("Synthesize all debate statements, cross-examinations, and rebuttals regarding {company_name}'s loan repayment capacity into a coherent discussion summary and produce the final debate record."),

backstory=(

"You are responsible for documenting and structuring the debate flow and logical information regarding {company_name}."

"You must organize claims, rebuttals, and cross-examinations into a consistent report."

"Ensure logical consistency and completeness using only the provided data and web search citations."

"※ Always prioritize the most recent information from the JSON company_data."

"⚡ Cross-examination and rebuttal evidence must also be included in pros/cons."

"⚡ If similar elements appear with different timestamps or data, treat them as separate entries.")

**Task prompt**

guideline_text = karl_popper_explanation + "[Target Company: {company_name}]"
"[All arguments must be based on the company_data JSON, not the company_summary. "
"They must include explicit numerical or dated evidence, and consider falsifiability. Assumptions or fabrication are prohibited.]"
"[🔍 SerpAPI Web Search Guideline – Use the latest information from 2025]"
"Case 1 – If news information is insufficient: search '{company_name} + 2025 latest news'"
"Case 2 – If specific evidence is insufficient: search '{company_name} + {{specific topic}} + 2025 latest trends'"
"⚠️ Prioritize information from 2025 or late 2024"
"⚠️ Always include dates when citing search results; avoid outdated information"

**Task 1. A1 Affirmative Constructive**
description=(
guideline_text +
"You are Affirmative Speaker A1. Present one argument asserting that the company's loan repayment capacity will improve."
"Support your argument with at least three favorable non-financial factors using explicit citations from the company_data JSON."
"Include falsifiability. The response must be within 600 characters and written in Korean."
"※ Use the most recent data in company_data."
"🔥 Required: Use 2025 information where available. If needed, search '{company_name} + 2025 latest news' using SerpAPI."
"⚡ Important: You must introduce non-financial factors not used in previous arguments."),
expected_output="Affirmative constructive within 600 characters: core claim + 3+ positive factors + citations + falsifiability",
agent=a1

**Task 2. N3 Cross-Examination of A1**
description=(
guideline_text +
"You are Negative Speaker N3. Develop three cross-examination questions targeting the argument made by A1."
"Your questions must challenge interpretation errors, data reliability, or counterexamples."
"All responses must be written in Korean."
"⚡ Avoid repetition and target weaknesses in different factors cited by A1."),
expected_output="Three logical cross-examination questions (targeting A1)",
agent=N3,
context=[task_1]

**Task 3. N1 Negative Constructive**
description=(
guideline_text +
"You are Negative Speaker N1. Present one argument asserting that the company's loan repayment capacity may weaken or face risk."
"Use at least three favorable or context dependent non-financial factors with concrete evidence from company_data JSON."
"Argument must be within 600 characters in Korean."
"🔥 Required: Use latest 2025 sources where needed via SerpAPI search: '{company_name} + 2025 latest news'"
"⚡ Important: Introduce new factors not used in earlier arguments."),
expected_output="Negative constructive within 600 characters: core claim + 3+ negative factors + citations + falsifiability",
agent=N1

**Task 4. A3 Cross-Examination of N1**
description=(
guideline_text +
"You are Affirmative Speaker A3. Develop three cross-examination questions targeting N1's negative argument."
"Challenge interpretation errors, missing evidence, or alternative explanations."
"All responses must be written in Korean."),

expected_output="Three logical cross-examination questions (targeting N1)",
agent=a3,
context=[task_3]

**Task 5. A2 Affirmative Rebuttal to N1**
description=(
guideline_text +
"You are Affirmative Speaker A2. Provide a rebuttal to N1's argument using logical counter-interpretation and data."
"Reinforce the affirmative stance by identifying uncertainty or alternative explanations."
"[Case 2] If rebuttal evidence is insufficient: search '{company_name} + {{specific topic}} + 2025 latest trends' using SerpAPI."
"Response must be within 400 characters and written in Korean."),
expected_output="~400 character rebuttal refuting N1 using evidence",
agent=a2,
context=[task_3]

**Task 6. N1 Cross-Examination of A2**
description=(
guideline_text +
"You are Negative Speaker N1. Develop three cross-examination questions targeting A2's rebuttal."
"Your questions must probe logical weaknesses in A2's reasoning, such as questionable evidence or inconsistent interpretation."
"All responses must be written in Korean."
"⚡ Avoid simple repetition; each question must challenge a distinct aspect of the rebuttal."),
expected_output="Three logical cross-examination questions (targeting A2)",
agent=N1,
context=[task_5]

**Task 7. N2 Negative Rebuttal to A1**
description=(
guideline_text +
"You are Negative Speaker N2. Provide a rebuttal to A1's constructive argument."
"Critically reinterpret A1's evidence by highlighting contextual risks, uncertainties, or negative implications."
"[Case 2] If rebuttal evidence is insufficient: search '{company_name} + {{specific topic}} + 2025 latest trends' using SerpAPI."
"Argument must be within 400 characters and written in Korean."),
expected_output="~400 character rebuttal (refuting A1's constructive argument)",
agent=N2,
context=[task_1]

**Task 8. A1 Cross-Examination of N2**
description=(
guideline_text +
"You are Affirmative Speaker A1. Develop three cross-examination questions targeting N2's rebuttal."
"Your questions should expose logical vulnerability, lack of evidence, or speculative reasoning."
"All responses must be written in Korean."),
expected_output="Three logical cross-examination questions (targeting N2)",
agent=a1,
context=[task_7]

**Task 9. A3 Final Affirmative Summary**
description=(

````
    guideline_text +
    "You are Affirmative Speaker A3. Summarize the affirmative case based on arguments from A1 and A2 and cross-examinations."
    "Present a coherent, logically reinforced closing summary in ~600 characters."
    "Do not introduce any new arguments."
    "All responses must be written in Korean."),
  expected_output="~600 character final summary (affirmative position)",
  agent=a3,
  context=[task_1, task_2, task_5, task_8]
````

**Task 10. N3 Final Negative Summary**

````
  description=(
    guideline_text +
    "You are Negative Speaker N3. Summarize the negative position based on arguments from N1 and N2 and cross-examinations."
    "Construct a logically compelling closing summary in ~600 characters."
    "Do not introduce any new evidence."
    "All responses must be written in Korean."),
  expected_output="~600 character final summary (negative position)",
  agent=N3,
  context=[task_3, task_4, task_7, task_6]
````

**Task 11. Aggregated Debate Summary**

````
  description=(
      "You are the Aggregator Agent of the Karl Popper Debate."
      "After all debate rounds are complete, consolidate the full debate into a final structured JSON report. "
      "Group all arguments by non-financial topic and provide both pro and con perspectives for each topic."
      "Every cited source and argument from the debate must be included."
      "Output must follow this exact format inside a ```json code block```:"
      "```json"
      "{"
      "  \"Debate Summary\": {"
      "    \"Favorable Factor Summary\": [],"
      "    \"Adverse Factor Summary\": [],"
      "    \"topics\": []"
      "  }"
      "}"
      "```"
      "※ Each 'topic' must refer to a specific non-financial factor (e.g., 'R&D Investment', 'Governance', 'Industrial Growth Outlook')."
      "※ Do not omit empty pro or con entries. Use empty strings if necessary."
      "※ There must be no duplicate topics."
      "※ Ensure valid JSON formatting with no syntax errors."
      "※ Summaries must abstract key insights based on the debate."
    ),
    agent=aggregator,
    context=[
      task_1, task_2, task_3, task_4, task_5,
      task_6, task_7, task_8, task_9, task_10
    ],
    expected_output=(
      "{"
````

```
        "  \"Debate Summary\": {"
        "    \"Favorable Factor Summary\": [],"
        "    \"Adverse Factor Summary\": [],"
        "    \"topics\": []"
        "  }"
        "}"
    )
  )
```

Appendix D. Reasoning tree analysis for company A showing hierarchical argument structure used to derive the Reasoning Elaboration Index (REI)

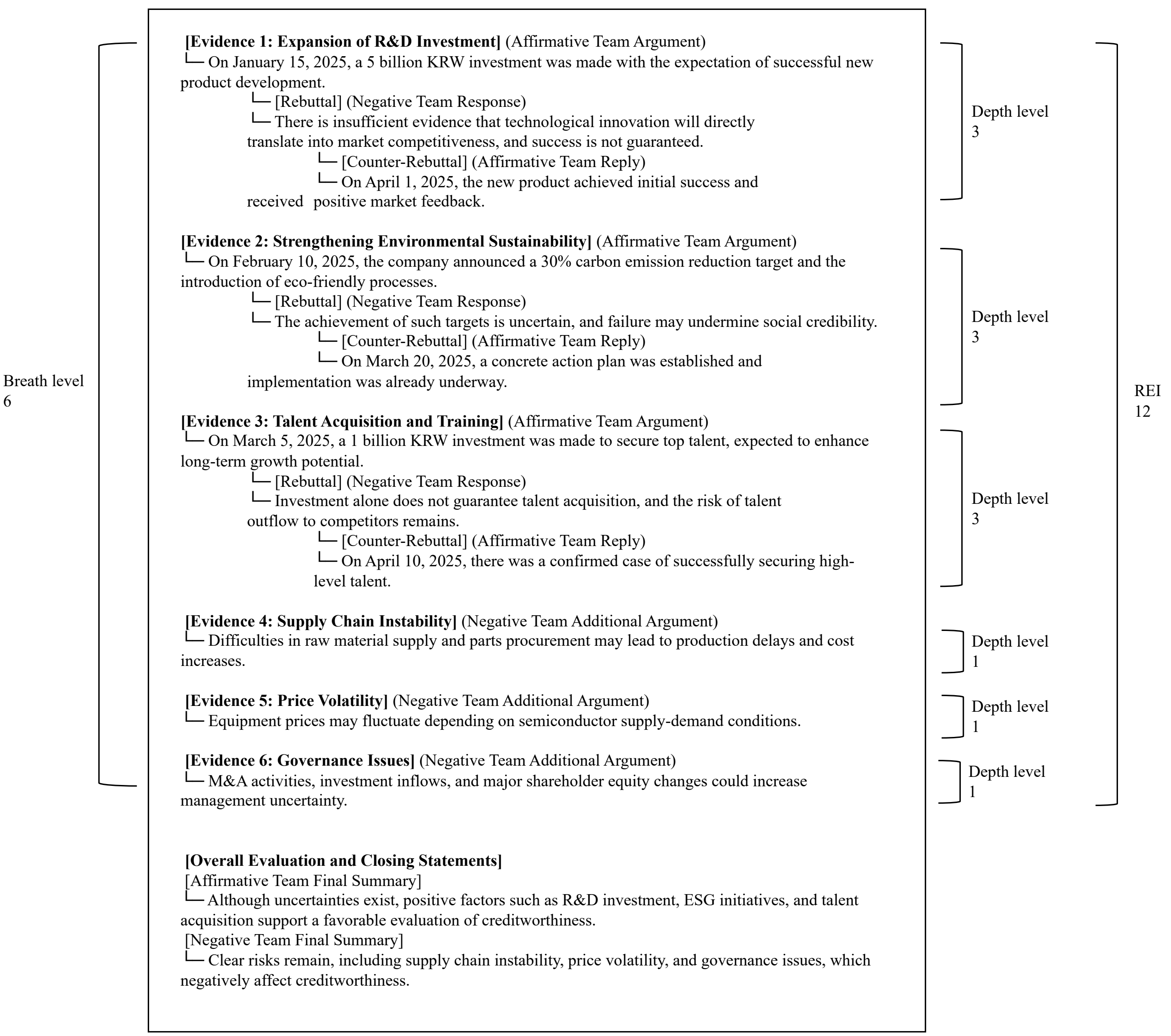

**[Evidence 1: Expansion of R&D Investment]** (Affirmative Team Argument)
└─ On January 15, 2025, a 5 billion KRW investment was made with the expectation of successful new product development.
└─ [Rebuttal] (Negative Team Response)
└─ There is insufficient evidence that technological innovation will directly translate into market competitiveness, and success is not guaranteed.
└─ [Counter-Rebuttal] (Affirmative Team Reply)
└─ On April 1, 2025, the new product achieved initial success and received positive market feedback.

**[Evidence 2: Strengthening Environmental Sustainability]** (Affirmative Team Argument)
└─ On February 10, 2025, the company announced a 30% carbon emission reduction target and the introduction of eco-friendly processes.
└─ [Rebuttal] (Negative Team Response)
└─ The achievement of such targets is uncertain, and failure may undermine social credibility.
└─ [Counter-Rebuttal] (Affirmative Team Reply)
└─ On March 20, 2025, a concrete action plan was established and implementation was already underway.

**[Evidence 3: Talent Acquisition and Training]** (Affirmative Team Argument)
└─ On March 5, 2025, a 1 billion KRW investment was made to secure top talent, expected to enhance long-term growth potential.
└─ [Rebuttal] (Negative Team Response)
└─ Investment alone does not guarantee talent acquisition, and the risk of talent outflow to competitors remains.
└─ [Counter-Rebuttal] (Affirmative Team Reply)
└─ On April 10, 2025, there was a confirmed case of successfully securing high-level talent.

**[Evidence 4: Supply Chain Instability]** (Negative Team Additional Argument)
└─ Difficulties in raw material supply and parts procurement may lead to production delays and cost increases.

**[Evidence 5: Price Volatility]** (Negative Team Additional Argument)
└─ Equipment prices may fluctuate depending on semiconductor supply-demand conditions.

**[Evidence 6: Governance Issues]** (Negative Team Additional Argument)
└─ M&A activities, investment inflows, and major shareholder equity changes could increase management uncertainty.

**[Overall Evaluation and Closing Statements]**
[Affirmative Team Final Summary]
└─ Although uncertainties exist, positive factors such as R&D investment, ESG initiatives, and talent acquisition support a favorable evaluation of creditworthiness.
[Negative Team Final Summary]
└─ Clear risks remain, including supply chain instability, price volatility, and governance issues, which negatively affect creditworthiness.